\documentclass[fleqn,10pt]{SelfArx} 

\usepackage[english]{babel} 

\usepackage{lipsum} 

\let\clsCenter\Center\let\clsendCenter\endCenter
\let\Center\undefined\let\endCenter\undefined
\usepackage{ragged2e}
\let\Center\clsCenter
\let\endCenter\clsendCenter

\usepackage{lineno}
\usepackage{url}

\usepackage[breaklinks]{hyperref}

\usepackage{caption}
\usepackage{subcaption}
\usepackage{amsmath,amssymb,amsfonts}
\usepackage{enumitem,mathtools}
\usepackage{algorithmic}
\usepackage{graphicx}
\usepackage{soul}
\modulolinenumbers[5]
\usepackage{multirow}

\usepackage{colortbl}
\definecolor{Gray}{gray}{0.85}
\definecolor{LightCyan}{rgb}{0.88,1,1}
\definecolor{mgreen}{rgb}{0.0,0.5,0.0}
\definecolor{brickred}{rgb}{0.8, 0.25, 0.33}
\newcolumntype{g}{>{\columncolor{Gray}}c}

\usepackage{ulem}

\usepackage{tabularx,multirow}

\definecolor{color1}{RGB}{0,0,90} 
\definecolor{color2}{RGB}{0,20,20} 

\usepackage{hyperref} 

\hypersetup{
    colorlinks=true,
    linkcolor=cyan, 
    filecolor=cyan,
    urlcolor=color2,
    citecolor=red,
    bookmarksopen=false,
    pdftitle={Title},
    pdfauthor={Author},
}

\JournalInfo{2020} 
\Archive{arXiv.2010.01746} 

\PaperTitle{Data-driven Operation of the Resilient Electric Grid: A Case of COVID-19} 

\Authors{
H. Noorazar\textsuperscript{1}, 
A. Srivastava\textsuperscript{1}*,
S. Pannala\textsuperscript{1},
K. S. Sajan\textsuperscript{1} 
}

\affiliation{\textsuperscript{1}\textit{School of Electrical Engineering and and Computer Science, Washington State University, Pullman, WA 99164, USA}}

\affiliation{*\textbf{Corresponding author}: anurag.k.srivastava@wsu.edu} 

\affiliation{*\textbf{Acknowledgement:} This work was partially supported 
by the U.S. Department of Energy UI-ASSIST Grant DE-IA0000025 
and the National Science Foundation award \#1840192. The views 
and opinions of authors expressed herein do not necessarily state 
or reflect those of the United States Government or any agency thereof. 
Technical support from Anshuman, Gowtham Kandaperumaland 
Jonah Davis is acknowledged} 

\newcommand{\keywordname}{Keywords} 

\Abstract{Electrical energy is a vital part of modern life, 
and expectations for grid resilience to allow a 
continuous and reliable energy supply has tremendously 
increased even during adverse events (e.g., 
Ukraine cyberattack, Hurricane Maria).  
The global pandemic COVID-19 has raised the electric 
energy reliability risk due to potential workforce 
disruptions, supply chain interruptions, and 
increased possible cybersecurity threats.  
Additionally, the pandemic introduces a significant degree of 
uncertainty to the grid operation in the presence 
of other challenges including aging power grids, high 
proliferation of distributed generation, market mechanism, and 
active distribution network. This situation increases the need 
for measures for the resiliency of power grids to 
mitigate the impact of the pandemic as well as 
simultaneous extreme events including cyberattacks and adverse weather events. Solutions to manage 
such an adverse scenario will be multi-fold: a) 
emergency planning and organizational support, 
b) following safety protocol, c) utilizing enhanced 
automation and sensing for situational awareness, 
and d) integration of advanced technologies and 
data points for ML-driven enhanced decision support. 
Enhanced digitalization and automation resulted in 
better network visibility at various levels, 
including generation, transmission, and distribution. 
These data or information can be employed to take 
advantage of advanced machine learning techniques 
for automation and increased power grid resilience. 
In this paper, 
we explore the resilience of power grids in the face of pandemics 
and discuss various machine learning tools that can be helpful
to augment human operators by: 
a) reviewing the impact of COVID-19 
on power grid operations and actions taken by 
operators/organizations to minimize the impact 
of COVID-19, and b) presenting  recently 
developed tools and concepts 
of machine learning and 
artificial intelligence that can be applied to 
increase the resiliency of power systems in normal 
and extreme scenarios such as the COVID-19 pandemic.}

\begin{document}

\maketitle 


\thispagestyle{empty} 


\section{Introduction} 

\addcontentsline{toc}{section}{Introduction} 
Electricity is essential for daily life, 
and the electric power industry is the 
backbone of the country's economy. 
Interruption in the flow of energy 
will lead to catastrophic impacts on 
other critical infrastructure, like 
the inability to manufacture essential 
needs and the hindrance of healthcare 
systems. The consequences are far-reaching 
and impactful. Consistent 
supply of electricity is a matter 
of national security: during the 
COVID-19 pandemic, ``the Department 
of Homeland Security listed power 
plants, dams, nuclear reactors, 
communications, and transportation 
systems as critical infrastructure.''~\cite{DHSCriticalinfra,tdworldsecutiry}.

\begin{figure*}[ht]
  \centering
  \captionsetup{justification=centering}
  \includegraphics[width=.7\textwidth]{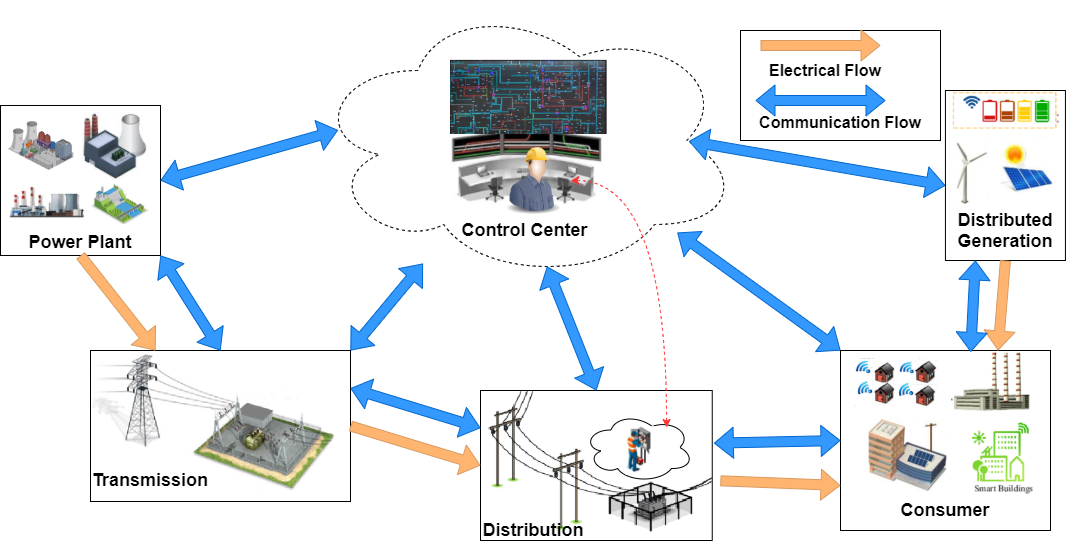}
  \caption{Overview of the Power system operation and control}
  \label{fig:PSO}
\end{figure*}

Power grids are actively generating and 
are continuously 
being monitored and maintained to provide
reliable energy for each customer.
The transmission ecosystem is under constant
monitoring to assess
the condition of the system and to execute
appropriate actions if needed.
Figure~\ref{fig:PSO} shows the operation
and control layout of existing power grids.
Distribution control room operators must monitor grid load and coordinate with field crews during routine maintenance and unexpected events~\cite{Adams2015} to restore power whenever abnormal circumstances cause outages.
\subsection{Background}
Having
robust contingency plans in the case of emergencies is vital. The
energy and utility sector in the U.S. is well planned
and designed to withstand disasters and
emergencies. Still, ``since 1980, there have been 246
weather and climate disasters exceeding \$1.6 trillion
in remediation''~\cite{Talley2020}. Contingency
planning has also improved and has been reshaped by 
pandemics of the past~\cite{eenewsContingency}.
Examples of such experiences are the Black Death, the 1918
Spanish Flu, the influenza pandemics of 1957 and 1968,
the SARS outbreak in the mid-2000s, and the 2009 H1N1 influenza.
Nevertheless, the novel situations caused by COVID-19,
with its fast spread and high transmissibility rate, have significantly
affected the utility industry. Most of the contingency
and backup plans are designed for physical assets and
do not consider the health of
employees~\cite{mcguirewoods}.
We emphasize the fact that most, 
if not all, of the literature studying the resilience of
power grid systems focuses its analysis on extreme weather
conditions or cyberattacks that threaten the 
physical components of a power grid. To date, extant literature on the human aspects of power grid resiliency is very poor. 
For example,
Jufri et al.~\cite{JUFRI20191049} explore definitions
of resiliency, grid exposure, and grid vulnerability 
in their review paper. 
All of the resources explored by~\cite{JUFRI20191049}
focus on extreme weather events as
low-probability high-impact events.
Building barriers and walls to protect 
physical assets against
flood is discussed in the literature and technical
reports~\cite{reportFlood, 6863387}. However, the
human side of power systems is taken for granted.
COVID-19 has impacted operators, 
forcing them to work
remotely or be quarantined, and consequently, utility
operations have been affected. Simmons and
Stiles~\cite{hstoday} and Terry~\cite{MarkTerry}
compare characteristics of COVID-19  and the 1918 Flu
and the potential problems that can occur today due to
coupled infrastructure, and authors from ~\cite{politico133888}
recounts painful lessons and what can be learned from
the 1918 Flu pandemic. 
In addition to the pandemics mentioned in
previous paragraphs, electrical disturbances caused
by weather have increased over the past
several decades~\cite{ChenResil}. It is estimated that power
outages have cost the United States economy \$18-\$33
billion between 2003 and
2012~\cite{ECONOMICWhiteHouse}. The average cost for
each weather event totals roughly \$150,000, and 50
events could happen every year, resulting in losses
of \$3 million annually~\cite{IBMresponse}. 
On a smaller scale, EPB Electric  Power in Chattanooga,
TN, estimated that the annual cost of power outages to
the community is approximately \$100 million.
By deploying automated switching and smart meters, they
saved about \$1.5 million in operational savings during
the  July 2012 wind storm~\cite{ElectricAdversarial}.
According to EPRI, ``severe droughts in recent years
have sharply reduced hydropower capacity in the West
and Southeast (U.S.) by
50\%''~\cite{EpriResilReport}. According to the survey
by IBM, proactive data-driven solutions can save a
significant amount each year. Natural disasters are
local (both in time and space); however, the COVID-19
pandemic spreads with people at high rates. Furthermore, it is
not yet known whether a person can get infected multiple 
times~\cite{CORONAVIRUSIndependent,theconversation}.
Preliminary studies suggest antibodies may last only a
few weeks or months~\cite{HeidtAmanda,liu2020prevalence}. If one can get infected more than once, then the
severity of the problem would be magnified. 
System
operators are highly skilled, and the expertise of an
experienced grid operator is significant. Hence, the
replacement of the workforce is challenging in case of
any loss.
\subsection{Motivation}
Due to the novelty of the pandemic and the lack of identifying the symptoms prematurely, operators of a control center became affected, and consequently, a power generation unit was shutdown~\cite{EnsuringEnergy}. Subsequently, all members of the electric power industry are actively updating their plans/guidelines in response to the ever-changing impacts of COVID-19. The Electricity Subsector Coordinating Council (ESCC) is releasing and updating its resource guides, including recommendations for decision-makers of industry~\cite{nercSpecialReport}, and the North American Reliability 
Corporation (NERC) is providing (and updating) its planning recommendations~\cite{Shaker6}. To cope with unforeseen events such as pandemics and/or other natural disasters, we need to utilize all the available tools. Eliminating or reducing manual interventions will ensure resilient and reliable grid operations during normal situations, in extreme natural events, and under scenarios such as the COVID-19 pandemic. The reduction of manual interventions will help utility field crew reduce their exposure to people who could be either symptomatic or asymptomatic carriers of the virus. These tools not only help in pandemic situations but also are helpful in general and under normal conditions. The power grid management has evolved with the invention and advances in computer technologies~\cite{PastPresentFuture} but is still behind and has not fully integrated newly developed research methods. Data analytics and machine learning (ML) tools can play a significant role in formulating effective solutions and strategies for managing natural events~\cite{IBMdataanalytics}. 
Moreover, as power systems become more complex, 
human operators must work in various situations 
and may encounter different types of 
emergencies~\cite{Yingkai2018}. 
More than two decades ago, in 1998, 
Bilke stated, ``trends in the electric utility 
industry have increased 
the chances and costs of errors by power 
system operators''~\cite{Bilke1998}. 
Using more sensors leads to the collection 
of more data. Consequently, 
the analysis of such vast data is beyond the cognitive capabilities of 
humans~\cite{Prostejovsky2019}, and more 
tools for assisting the operators are
needed~\cite{Guttromson2007,Prostejovsky2019, Yuanjun2018}. 
Innovative technologies can improve operator 
knowledge of the state of the power system 
and thus make more efficient and reliable 
operation possible~\cite{TheFutureOfGrid}. 
Decentralization of power systems, shifting 
towards smart grids and consequently more 
complicated systems, and a scarcity of 
expert operators all mean new technologies 
and automation must assist human operators 
in their larger roles. Not only will machine 
learning be essential on the micro-level 
to perform some of the system operators' 
tasks but, on the macro level, new 
technologies are needed for decision-makers 
to make better informed decisions in the 
case of disasters~\cite{Talley2020}.
\subsection{Contributions}
Most of the recent COVID-19-related papers and reports have
focused on statistics in the reduction of power usage
and change of demand patterns~\cite{Sahand19,Barooah19}.
Some of these papers rushed to analyze the change of
pattern of power consumption. Barooah and Duzgun.~\cite{Barooah19}
claims COVID-19 does not solely cause the change in
demand pattern. While these COVID-19-related papers
report on impacts on load-demand, economy, and
implement strategies to deal with difficulties of the
pandemic~\cite{Paaso}, the main goal of this paper is
to recommend technical strategies that are not
integrated into the industry to date, and these
techniques will have a far-reaching impact if the industry
adopts them. Increasing the resiliency of power grids
helps utility companies to deal with situations such as
COVID-19 along with other natural events.  Resiliency \textit{usually} is defined under
high-impact low-probability weather events that impact
\textit{physical assets} of power grids. The COVID-19 pandemic has
affected power grids indirectly by impacting
operators. It is shown~\cite{Kraemer493} that early
restrictions on travel are effective, and late
restrictions (when the outbreak is widespread) are less
effective. Thus, implementing statistical learning
approaches enhances the resiliency of microgrids by
reducing the number of onsite operators.

Winter began in the Northern Hemisphere, and the importance of power
systems' resiliency cannot be overstated. In 2020, We
have already seen a flood on May 19th and
20th, in Midland County, Michigan~\cite{MichFlood},
as well as a Cyclone hurricane in India and Bangladesh,
on May 20, present adverse impacts on the power
transmission system~\cite{NYtimesbangladesh}. The scope
of our paper does not consider an analysis of natural
disasters. While planning, operating, and restoring
strategies in response to natural disasters are
discussed in a typical review paper (first three rows
of Table~\ref{tab:HILPtable}), we only discuss the
methods relevant to pandemic scenarios (last two rows
of Table~\ref{tab:HILPtable}).These models increase the
resiliency of the power grid by decreasing the
dependency on operators and increasing the automation.
These methods significantly improve the non-physical
problems that have mostly been ignored in the
past. Reference~\cite{Kandaperumal90}  reviews some of
the ideas related to human resource resilience.
\begin{table}[ht]
\centering
 \caption{High impact, low probability events} \label{tab:HILPtable}
\begin{tabular}{ll}
\hline
\rowcolor{Gray} Classification of Events & Examples \\ 
\hline
Physical, manmade & Riots, vandalism, war\\
 & ~\cite{7395278,8612343,4448770,national2012terrorism} \\ 

Physical, natural & Hurricane, earthquake, storms\\
& \cite{AMIRIOUN2019716,WangChen,MohamedChen}\\ 

Cyber & Phishing, data injection \\
& 
~\cite{8625421,HinkAdhikari,ZhangCSI,Ahmed2018Cyber}\\ 
Non-physical, natural & COVID-19, Spanish flu \\
\hline    
\end{tabular}
\end{table}

The originality and motivation for our paper 
compared to other review 
papers lies in the fact that we 
focus on reviewing challenges of 
power grids under pandemics and offer potential approaches 
to counteract them.
Thus, comparing different 
ML-based models (e.g. performance of the models)
against each other is beyond scope of this paper. In this paper 
we provide an overview of data-driven methods that leads to the
augmentation of human roles in operating power grids, which consequently
will be beneficial in cases such as the
COVID-19 pandemic.
Contributions of this paper include:
\begin{enumerate}
    \item Analyzed impact of COVID-19 on power 
    grid operations and actions being taken 
    by operators/organizations to minimize 
    the impact of COVID-19.
    \item Analyzed concept of resiliency and key differences with other existing concepts for the power grid
    \item Reviewed some of the recent ML-based methodologies 
    developed for increasing power systems' resiliency
    that are relevant under the pandemic 
    scenario and in general.
    \item Suggested integrating some of the recently 
    developed tools or concepts in ML and AI with 
    industry practices. These approaches can increase 
    the resiliency of power systems in general and 
    in extreme scenarios such as the COVID-19 pandemic.
    \item Presented an example case study for COVID-19 with the real-time resiliency management tool (RT-RMT)  developed by our team that uses 
    real-time data for optimal crew routing and 
    restoration process after a power outage during the COVID-19 pandemic.
\end{enumerate}

The rest of the paper is organized as follows. 
In Sec.~\ref{ExistingOperation2}, a quick review of existing
power grid operations is presented. Next, in Sec.
~\ref{COVIDonGrid3} impact of COVID-19 pandemic on power grid
operations is analyzed. Afterward, Sec.~\ref{CovidSolution4}
investigates possible and employed 
practices to manage and combat the
pandemic effects in normal operations, 
followed by ML applications for increasing 
resiliency of power systems in the future. 
In Sec.~\ref{futureDir}, we discuss some
research directions that can be explored further
in the future. Finally, we
present our conclusions in Sec.~\ref{conclusionSec}.
\section{Existing power grid operational practice}\label{ExistingOperation2}
Power
systems require a level of centralized planning and
operation to ensure system reliability. 
\subsection{Normal operation} 
The electric power system is built to handle periodic equipment failures, primarily by isolating faulty zones whenever any problems occur. The control room is the brain of a power grid, responsible for the reliablity and security of the network. These operations in the power system are a combination of automated control and actions that require direct human intervention. System operators at control centers carry out many of these centralized tasks, including short-term monitoring, analysis, and control.
\subsubsection{SCADA: data acquisition and control}
Supervisory control and data acquisition
(SCADA) helps in achieving situational 
awareness and effective decision making. SCADA is a system of
software and hardware elements
responsible for gathering, monitoring, and processing real-time data. 
SCADA acquires measurement information
related to voltage, current, 
and frequency, circuit breakers status,
monitoring active and
reactive power flow, etc. 
Collecting, processing, and
displaying real-time data allows
operators to analyze the 
data via human-machine interface
and enables them to take proper actions
remotely from the control center.
Currently, the presence of operators in
 control centers is necessary 
24 hours a day. 
Operators utilize the energy management 
system and distribution management 
system to monitor the state 
of the system such as the balance 
between generation and demand, fault signals, thermal 
loading of feeders, etc.

\subsubsection{Control center: situational awareness and decision making}

Situational awareness means being constantly aware of the state of power station or grid conditions. It is a significant contributor to successful control room operations. Observing the early signs of the system approaching a vulnerable state requires a wide-area view and analytics to recognize conditions. Situational awareness is vital to ensuring coordinated responses among operators within large organizations~\cite{situationroom}. Situational awareness gained through SCADA and utilizing sensors such as a smart meter, phasor measurement unit (PMU), and data acquired through the communication system provides an opportunity to develop continuous monitoring of the system\cite{Ahmed2018Cyber}.

Table~\ref{tab:challengeTablePG} presents current challenges faced by electrical power systems where 
ML approaches can alleviate these problems and help operators in different tasks with decision support tools. Research methodologies have been developed to battle these challenges but are not yet integrated into the systems. Primarily, these approaches will reduce dependency on human operators, which is helpful in situations such as the COVID-19 pandemic.

{{\small
\begin{table*}[ht!]
\centering
\caption{
Challenges for the power grid operation and control in extreme events.} \label{tab:challengeTablePG}
\begin{tabular}{ c c c c }
\hline
\rowcolor{Gray} Field & Challenge & Opportunity (Developed Methods) & Refs. \\
\hline

\multirow{1.5}{6em}{Control Center} &
\multirow{2.3}{15em}{Fault detection due to incomplete and conflicting alarms.}  & 
\multirow{2.3}{15em}{Data-driven based on mixed integer linear programming.} & \multirow{2}{2em}{\cite{Jiang88}} \\ \\ \\ 
& \multirow{1}{15em}{Increasing number of measurement devices increases the number of alarms causing analysis and situational awareness more challengiing.} & \multirow{1}{15em}{Real-time event detection using synchrophasor data, and ensemble methodology that includes maximum likelihood estimation, DBSCAN, and decision trees.} & \cite{Pandey9072390} \\ \\ \\ \\ \\
& \multirow{3}{15em}{Fault section estimation.} &  
\multirow{3}{15em}{Neural networks}  & \multirow{2.5}{4em}{\cite{Cardoso13,Wang3,reddy2008neural,Ghanizadeh68}} \\ \\ \\

& \multirow{.1}{15em}{Classifying 
the observed anomalies of 
instrument transformers into
different types of malfunctions,
failures, or degradation.} &
\multirow{3}{15em}{A pipeline
consisting of three steps; maximum
likelihood estimation, DBSCAN and a
decision logic diagram.}  &
\cite{Cui89} \\ \\ \\ \\ \\
\hline
\multirow{1.5}{6em}{Office Staff} & \multirow{1.5}{15em}{Classifying the customer ticket texts/calls.} & \multirow{1.5}{15em}{Natural language processing, text analytics, recurrent neural networks.} & \cite{Rudin2014} \\ \\ \\ 
\hline 

\multirow{1.5}{6em}{Transmission protection systems} & \multirow{1.5}{15em}{Detecting root cause 
of failures in transmission protection systems. 
Failure to do so causes propagation of 
faults, consequently resulting in multiple 
conflicting alarms. 
} & \multirow{1.5}{15em}{Anomaly detection using an ensemble of ML approaches.} & \cite{Gholami2019} \\ \\ \\ \\ \\ \\ \\

\multirow{1.5}{6em}{System Protection - 
Cyber Security
} & \multirow{1.5}{15em}{Monitoring and detecting malicious activity in transmission protection systems. 
} & \multirow{1.5}{15em}{Anomaly detection 
pipeline that includes LSTM networks, 
semi-supervised deep learning algorithms, 
and ridge regression.} & \cite{Ahmed2018Cyber} \\ \\ \\
& \multirow{5}{15em}{Classifying malicious data and possible cyberattacks.} &  
\multirow{5}{15em}{Several ML algorithms such as random forests and support vector machine.}  &  \multirow{4}{2em}{\cite{HinkAdhikari}} \\ \\ \\ \\

& \multirow{3}{15em}{Detecting anomalies and false data injection.} & \multirow{3}{15em}{Gaussian mixture model is used to solve the problem.}  & \multirow{2}{2em}{\cite{ForoutanDetection}} \\ \\ \\

& \multirow{3}{15em}{Detecting intrusions in smart grids.} &  \multirow{3}{15em}{Support vector machine and artificial immune system.}  & \multirow{2}{2em}{\cite{ZhangCSI}}\\ \\ \\

\hline
\end{tabular}
\end{table*}
}}
\subsection{Contingent operation: power outage and restoration}
Power outage, restoration, and assessment
management plan vary between expected 
contingency and emergency operations. 
The following section will discuss the 
preparation for events and restoration 
efforts during these two modes of operations.
\subsubsection{Planning and training for normal expected contingency}
In normal expected contingency operation, 
utilities prepare for all sorts of 
circumstances ranging from small 
storms, winter snow and ice storms, cyclones, 
insufficient resources of generation, 
shortage of fuel reserves, accidents, cyberattacks, etc.
There are several activities that utilities perform during regular operation to prepare for events in advance: 
\begin{itemize}
    \item Operators and field crews perform exercises and drills to prepare for various incidents. 
    \item Modernizing physical infrastructure to make it less vulnerable to physical or cyber events.
    \item Utilities complete contingency planning to ensure they can maintain supply even if one or more system components go offline.
    \item Utilities perform regular vegetation management, which includes the cutting or trimming of plants, bushes, and other foliage that could be too near electric infrastructure, preventing potential damage to equipment during storms.
    \item Utilities regularly examine resources, noting deterioration and required repairs or replacements.
\end{itemize}
\subsubsection{Planning and Resilient Operation for Emergency}
During emergency operation
mode, utilities coordinate with
control room operators and
crews to re-establish power
supplies as fast as possible.
If the event is known
beforehand, then utility
planning will be in two
phases, pre-event and
post-event.\\
\textbf{Pre-event process:}
\begin{itemize}
    \item Utilities create different leads for different functions (e.g., damage, restoration, vegetation management, overall communications).
    \item Utilities review critical infrastructure in the area.
    \item Utilities identify resources, including crews, backup generators, and other equipment, as well as mutual assistance available to respond to emergencies.
    \item Utilities employ advanced tools to anticipate extreme events and to prepare the system to have minimal impact.
\end{itemize}
\textbf{Post-event process:}
\begin{itemize}
    \item Utilities perform a destruction estimation of lines and substations.
    \item Utilities eliminate all hazardous situations, such as downed live wires or other potentially life-threatening situations.
    \item Utilities restore power plant and transmission lines that carry maximum power to the distribution system whenever damaged.
    \item Utilities prioritize restoring power to critical infrastructure such as hospitals, police, and fire stations before individual homes and small businesses.
    \item Utilities employ advanced resiliency tools to optimize and recover after extreme events.
\end{itemize}
\section{Impact of COVID-19 on the power 
grid operation}\label{COVIDonGrid3}
Edison Electric Institute in their 2020 report~\cite{eei}, 
states that ``it is predicted that a large 
percentage of a company's
employees (up to 40\%) could be out sick, 
quarantined, 
or might stay home to care for sick family members.'' Moreover, certain nuclear
facilities have operated marginally at low 
capacity due to remote operations and the challenges 
presented by delayed maintenance work, all of which
impacts the timely operation of such facilities. Furthermore,
COVID-19 is expected to delay project developments and
impact renewable auctions due to supply chain delays. The pandemic
has forced the NERC to delay implementation of several reliability
standards~\cite{nercInter, UNITEDSTATES}, some of which are
cybersecurity measures. The pandemic has 
forced utilities to operate
with skeleton crews, providing more opportunities for
cyberattackers~\cite{nercSpecialReport}. 
Three infected IT security
engineers at a nuclear power plant were quarantined for 14
days~\cite{EnsuringEnergy}, increasing the threat to normal operations. 
State of the art technologies in AI and ML can aid 
in reducing humans' direct involvement with 
oversight, maintenance, and failure
detection~\cite{Gholami2019}. Protection systems are 
always of the highest priority in the energy industry.

For increased security against
cyberattacks, extra measurements must be taken
(see~\cite{Ahmed2018Cyber}).

\begin{figure*}[ht]
\centering
    \begin{subfigure}[b]{0.4\textwidth}
        \includegraphics[width=1.2\textwidth]{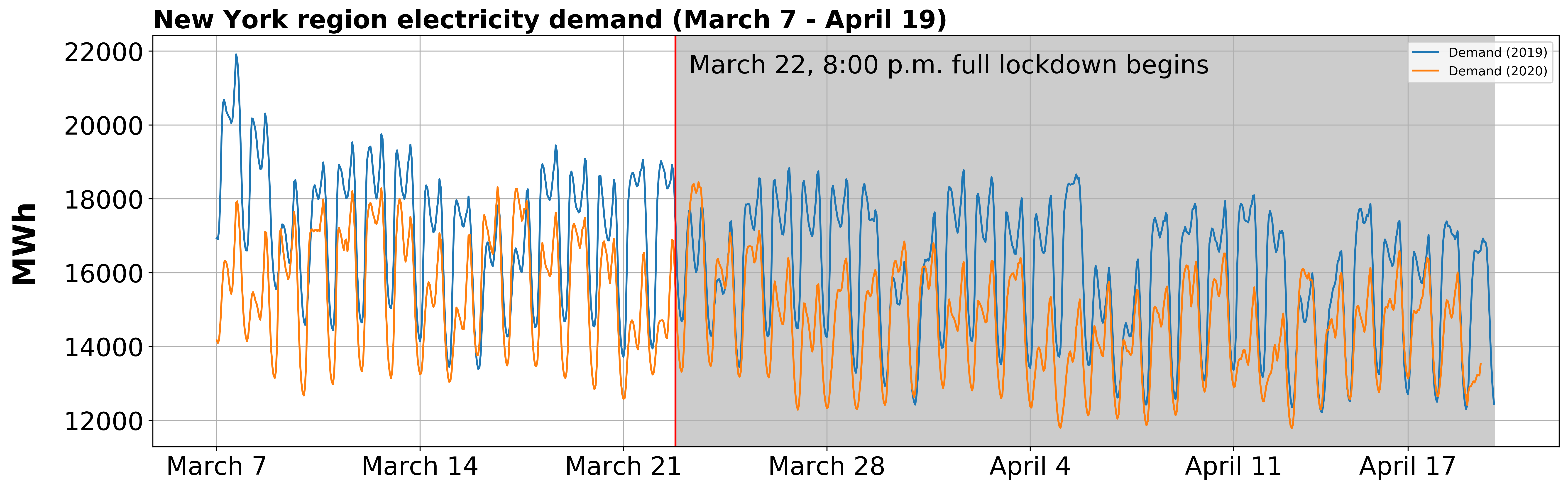}
        \caption{New York's demand pattern.}
        \label{fig:NYDemand}
    \end{subfigure}\qquad \qquad
    \vspace{.1in}
    ~
    \begin{subfigure}[b]{0.4\textwidth}
        \includegraphics[width=1.2\textwidth]{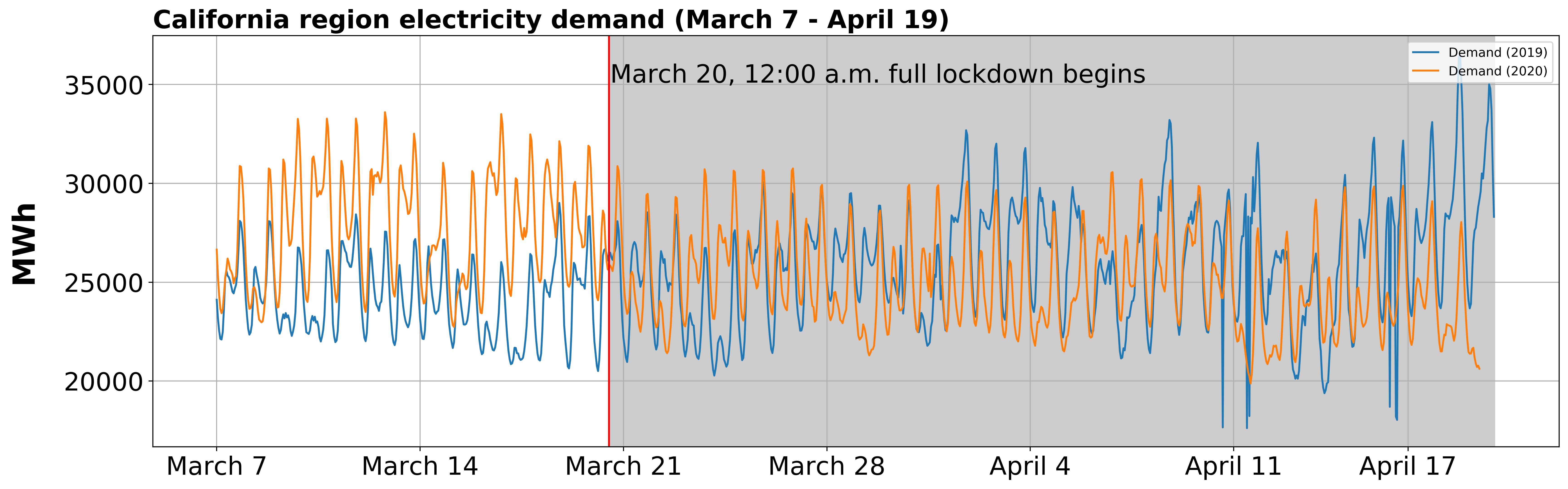}
        \caption{California's demand pattern.}
        \label{fig:CaliDemand}
    \end{subfigure}
    \begin{subfigure}[b]{0.4\textwidth}
        \includegraphics[width=1.2\textwidth]{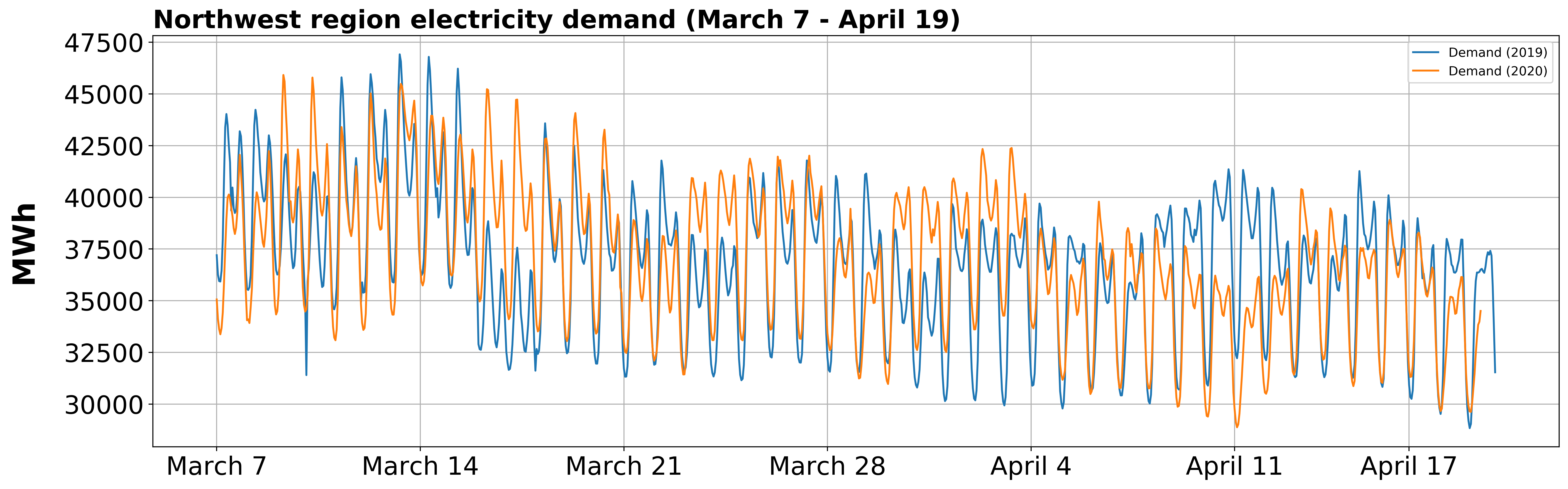}
        \caption{Northwest region's demand pattern.}
        \label{fig:NWDemand}
    \end{subfigure}\qquad \qquad
    \begin{subfigure}[b]{0.4\textwidth}
        \includegraphics[width=1.2\textwidth]{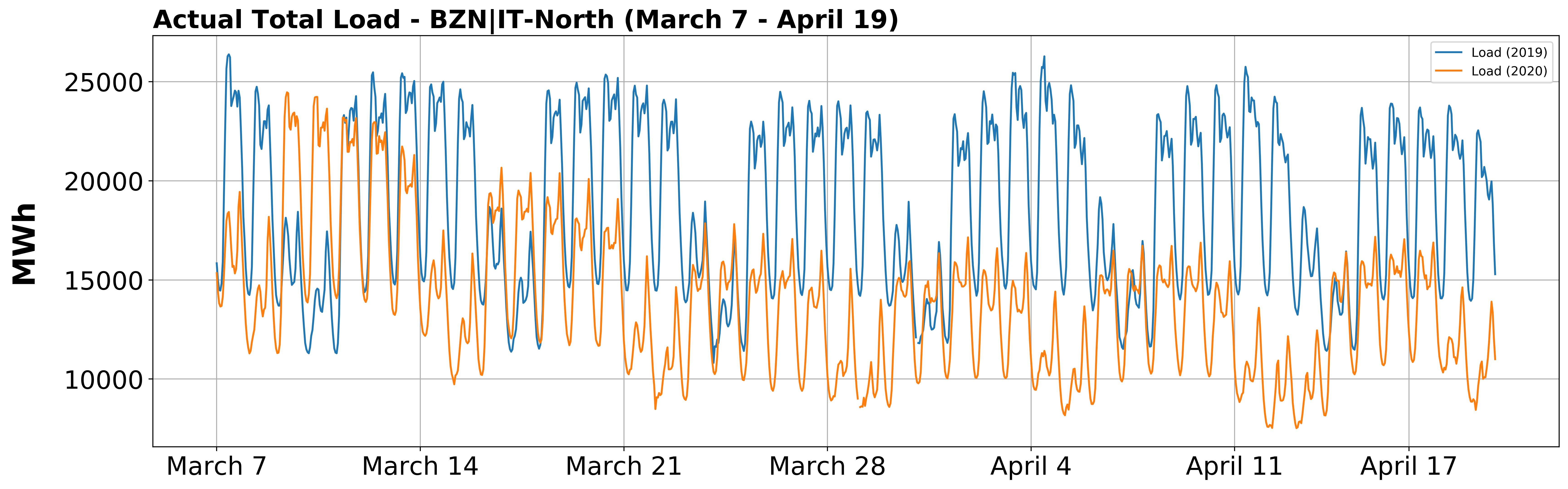}
        \caption{Northern Italy's load pattern.}
        \label{fig:NItalyLoad}
    \end{subfigure}
    \caption{COVID-19 impact on demand behavior in three regions in the US and in Northern Italy.}
\label{fig:DemandPatternPlots}
\end{figure*}
COVID-19 impacted power demand, mostly when lockdown periods went into effect. Electric Power Research Institute (EPRI) reported the change of pattern of load in Italy, Spain, and the U.S.~\cite{EPRIReport}. Load impacts in the U.S. were also reported in~\cite{spglobal,utilitydive,hstoday}.

Figure~\ref{fig:DemandPatternPlots} compares the
power consumption in 2020 and 2019 in various
regions of the U.S. and Italy when the pandemic
became widespread, and lockdowns began. The
data for the U.S. is obtained from Energy
Information. The data for Italy is obtained from
ENTSO-E Transparency~\cite{ENTSOTransparency}.
The lockdown started on March
19~\cite{CaliExOrd} in the state of California.
The demand in California was higher before the
lockdown began in 2020 compared to 2019.
However, after the lockdown, it dropped below
that of 2019 by a small amount. The Northwest
region includes several states; therefore, there
was no uniform date for the beginning of
lockdowns in the region, and no clear change of 
pattern is seen in the Northwest.
In the state of New York, lockdown started on
March 22~\cite{CumoExOrd}. Here, demand in
2020 was less than demand in 2019, even
before the lockdown period began.
This difference might indicate that other
factors (beyond COVID-19) might be involved
in the difference between consumption loads of
2020 and 2019. 

Neural networks can be used for
short-term load forecasting. Badri et al.
~\cite{BADRI20121883} compare neural networks
against fuzzy logic in the context of short-term
load forecasting and reports on superiority of
neural networks. Additionally, Dudek~\cite{DUDEK201664}
compares different neural networks against each
other for the same goal. The authors have to admit a 
rapid change in behavior of power consumption
as a result of lockdowns. Thus, a lack of sufficient
data for ML-approaches will be a challenge.

Ruan et al.~\cite{ruan2020cross}
have made an ensemble model to predict
electricity consumption in NYISO in 2020 if
COVID-19 were absent. They trained 800 neural
network models, selecting the top 25\% models
with the lowest error. For certain selected days in
April, they showed that electricity consumption
was much lower than predicted by the
model. Table~\ref{tab:demandCompare2019}
presents reduction of power consumption in
different countries in 2020 compared to 2019. 
The reduction of electricity usage translates
into a reduction of the system's resilience.
Furthermore, most of the reduction of demand
results from a reduction of electricity usage by
large segments such as industrial and commercial
organizations. These customers form a large segment
of normal electricity demand, and their lack of
stability translates into a reduction of
stability in power grid operations.

In the U.K., 
The consumption
of the transmission system kept falling week
after week; ``compared to the week of March 9th,
total electricity demand fell by 13\% in the
week of March 23rd, by 14\% in the week of March
30th, and by 24\% in the week starting April
6th.''~\cite{COVIDUKD}. Europe went through a
significant reduction (27\%) during lockdowns.
Even after the easing of lockdowns in June, the
demand was still 10\% less than that of
2019~\cite{EuropeEase}.

{\small
\begin{table*}[ht]
\centering
\setlength{\tabcolsep}{4pt}
 \caption{{\color{black}{Comparison of demand in different regions and time periods.}}}\label{tab:demandCompare2019}
 \begin{tabular}{ l l l l }
\hline
\rowcolor{Gray}
Country & Demand reduction & Time Window & Refs. \\
\hline
\multirow{1.5}{6em}{USA} & 
\multirow{1.5}{10em}{5.4\%} &
\multirow{2.3}{14em}{Week of April 6, 2020 compared to the same time in 2019} & 
\multirow{1.5}{3em}{\cite{weforum}} \\ \\ \\
\hline

\multirow{1.5}{6em}{NYISO (USA)} & 
\multirow{1.5}{10em}{9\% reduction in peak demand} &
\multirow{2.3}{14em}{March 2020 compared to 5-year average values} & 
\multirow{1.5}{3em}{\cite{Paaso}} \\ \\ \\
\hline

\multirow{1.5}{6em}{CAISO (USA)} & 
\multirow{1.5}{10em}{5\%-8\% load reduction in weekdays} &
\multirow{2.3}{14em}{Since statewide shutdown on March 20 compared to 2019.} & 
\multirow{1.5}{3em}{\cite{CAISOreduc}} \\ \\ \\
\hline

\multirow{1.5}{6em}{Europe} &
\multirow{1.5}{10em}{27\%} & 
\multirow{1.5}{14em} {The first week in February to the end of March in 2020 compared to 2019.} & 
\cite{RenewablesEurope} \\ \\ \\
\hline 

\multirow{1.5}{8em}{France, Germany, \\ Italy, Spain, and Netherlands.} &
\multirow{1.5}{10em}{10\%} & 
\multirow{1.5}{14em} {The first week in February to the end of March in 2020 compared to 2019.} &
\cite{RenewablesEurope} \\ \\ \\
\hline 

\multirow{1.5}{8em}{India} &
\multirow{1.5}{10em}{30\%} & 
\multirow{1.5}{14em} {Since full national lockdown on March 24.} &
\cite{weforumIndia} \\ \\
\hline 
\end{tabular}
\end{table*}
}
Some interesting results, such 
as the impact of social distancing on 
electricity consumption  can be 
found in~\cite{ruan2020cross}.

To prevent the spread of the virus, operators 
and technicians have had to change their normal
operations. Different solutions have been proposed by experts. The followings
are some examples of the pandemic's impacts and
potential mitigations to consider.
\begin{itemize}
    \item Minimizing the number of onsite staff and working remotely if possible.
    \item Using main and backup control rooms at
          different times and/or with separate
          operators.
    \item Regularly sterilizing control center facilities
     during the day.
    \item Changing shift duration and cycles.
    \item Limiting common equipment usage.
    \item Cancelling face-to-face meetings and direct interactions.
    \item Limiting the number of field technicians in each vehicle and ensuring technicians carry personal protective equipment.
    \item Closing all non-critical common areas, such as exercise rooms or even cafeterias. 
\end{itemize}
\section{Solutions for minimizing impact of COVID-19}\label{CovidSolution4}
The COVID-19 pandemic has changed usual operations at different levels. 
Recently, the Federal Energy 
Regulatory Commission (FERC) and NERC
produced industry guidelines to reduce the 
impact of COVID-19~\cite{FERCNERCGuid}, and
a comprehensive guide is given by~\cite{AssessingMitigating}.
\subsection{Actions Being Taken by Grid Operators}
\begin{figure*}[ht]
  \centering
  \captionsetup{justification=centering}
  \includegraphics[width=.7\textwidth]{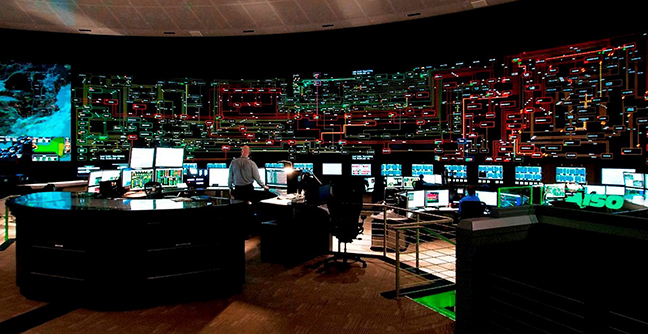}
  \caption{Control center normal operations are impacted by COVID-19. Power systems are operating with minimal number of staff onsite.}
  \label{fig:OpAct}
\end{figure*}

Operators are taking different actions
to minimize the spread and impact of SARS-CoV-2
while attending their jobs and maintaining  
the power stations. Below is a list of some 
of these strategies.
\begin{itemize}
\item Back-up control centers are activated, key workers are isolated, and deep cleaning
has become routine~\cite{AnnouncementRespo}.

\item Food and other supplies are stored at the generating station (Utah’s Intermountain Power Plant)~\cite{latimespandemic}.
\item Operators are preparing for the worst-case scenario; ``we
    could operate the plant with a skeleton crew
    for an extended period of time'' (Utah’s Intermountain Power Plant).
An example is shown in Fig.~\ref{fig:OpAct} 
provided by~\cite{eenewsNetSkeleton}, 
demonstrating New York's power grid 
control operation working with only essential operators.
\item Employees are telecommuting. (Southern California Edison, NYPA)
\item Rotating personnel between power line repair
teams is kept to a minimum (Southern California Edison)~\cite{latimespandemic}.
\item Non-critical planned maintenance is being postponed (Southern California Edison).
\item Critical workers are isolated (American Gas Assn.). In fact, ``some sequestration is underway in certain areas, with employees and contractors living on-site at power plants and other facilities. It is critical that sequestered employees who are in close quarters be tested before and during sequestration''~\cite{EnsuringEnergy}. 
\item Regular health screenings are being performed, as well as 
deep cleanings of control rooms to keep 
the coronavirus out (NYPA)~\cite{wiredcoronavirus}.
    \item Splitting control center operators between the main and backup control center to limit contact (PJM)~\cite{wiredcoronavirus}.
\end{itemize}
\subsection{Enabling grid resiliency}
Other than the usual operational 
paradigm to focus on reliability, security,
restoration, and emergency planning, energy suppliers need 
to invest in resiliency to bounce back from
extreme events. 
Resiliency is a multidimensional concept.
Jufri et al.~\cite{JUFRI20191049} 
state that resiliency 
``involves multidisciplinary knowledge, 
such as power
system, civil and structure, 
geography, computer
science, probability, and meteorology and
climatology.'' 
Consequently, several definitions are available. 
Even for one aspect of resiliency, several definitions 
have been put forward, and there is no consensus
in a given community. Definitions for
a given aspect are not mutually 
exclusive and may overlap~\cite{RePEc}.
Table~\ref{tab:ResiliencyDefTbl} 
provides some of the definitions 
of resiliency from different 
perspectives and/or 
in different domains, as well as for some 
closely related topics.  
For more details please see~\cite{Bie06,RePEc}.
\begin{table*}[ht]
\centering
 \caption{Definitions of resilience and other coupled/interrelated concepts.} \label{tab:ResiliencyDefTbl}
 \begin{tabular}{ l l l l }
\hline
\rowcolor{Gray}
Concept & Definition & Refs. \\
\hline
\multirow{1.5}{10em}{Resiliency} & \multirow{2}{24em}{Ability to harden 
the power system
against—and quickly recover from high-impact,
low-frequency events. Resiliency consists of
damage prevention, system recovery, and
survivability.
}  & 
\multirow{2}{2em}{\cite{EPRIResilDef} } \\ \\ \\ \\ \\

\multirow{1.5}{10em}{Damage prevention \\(as a part of resiliency)} & \multirow{1.5}{24em}{Damage prevention is
self-explanatory. It is related to predicting
and hardening the power systems against
high-impact, low-frequency events.
} & \cite{EPRIResilDef} \\ \\ \\ \\
\multirow{1.5}{10em}{System recovery 
\\(as a part of resiliency)} & 
\multirow{1.5}{24em}{Going back to normal
operations after occurrence of high-impact,
low-frequency events.
} & \cite{EPRIResilDef} \\ \\ \\ \\
\multirow{1.5}{6em}{Survivability
(as a part of resiliency)} & 
\multirow{1.5}{24em}{Ability to maintain some basic level of power
to consumers when complete access to their
normal power sources is not possible.} & ~\cite{EPRIResilDef} \\ \\ \\
\hline
\multirow{1.5}{10em}{Resiliency} 
& \multirow{2.5}{24em}{``measure of a system's 
 ability to absorb continuous and unpredictable 
 change and still maintain its vital functions.''} 
& \cite{RePEc} \\ \\ \\ \\
\multirow{1.5}{10em}{Resiliency in organizational domains} & \multirow{1.5}{24em}{``ability of
an organization to absorb strain and improve
functioning despite the presence of
adversity.''} 
& \cite{RePEc} \\ \\ \\
\hline
\multirow{1.5}{6em}{Reliability} & \multirow{1.5}{24em}
{``the degree to which the performance of the elements of
[the electrical] system results in power being delivered to consumers within accepted
standards and in the amount desired.''} 
& \cite{hirst2000bulk} \\ \\ \\ \\
\hline
\multirow{1.5}{10em}{Flexibility} & \multirow{2.5}{24em}{``the inherent capability to modify a current direction to accommodate and successfully  adapt to changes in the environment.''} 
& \cite{RFEHUSDAL} \\ \\ \\
\hline
\end{tabular}
\end{table*}
As mentioned before, most of 
the effort on resiliency is focused 
on physical assets. 
In this paper, however, 
we are dealing with disease, which
adds a different dimension to 
the problem: health. 
Consequently, the organizational domain 
becomes relevant, and a definition of resiliency
for this aspect is presented in Table~\ref{tab:ResiliencyDefTbl}.
By implementing ML approaches, the
resiliency of a power system--as a
group/society/organization--is increased as organizational responsiveness is
shifted from humans to machines.

\begin{figure}[ht]
  \centering
  \captionsetup{justification=centering}
  \includegraphics[width=.5\textwidth]{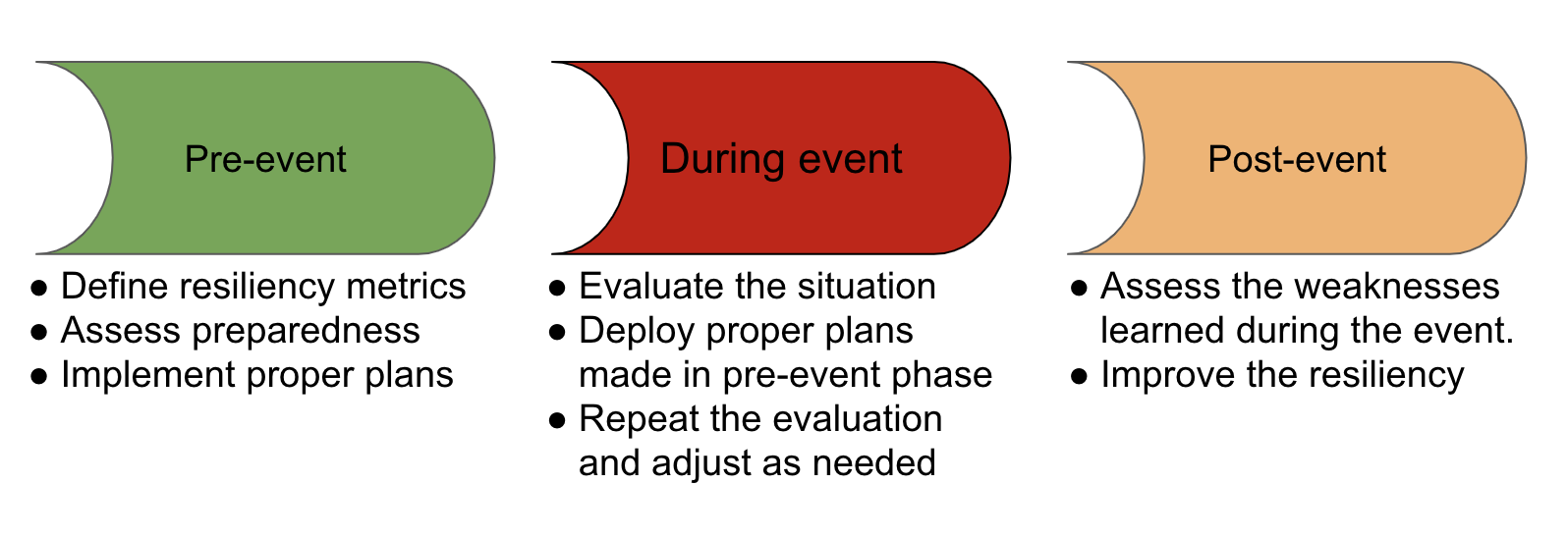}
  \caption{Multi-temporal resilience framework}
  \label{fig:temporalResil}
\end{figure}

In general, the resiliency of a 
power grid is defined as ``the ability 
to withstand and fast
recovery from deliberate attacks, 
accidents, or naturally
occurring threats or incidents''~\cite{PPD21}.
For a system to be resilient, it needs to have a
robust damage prevention plan in place, and a
system designed for adaptation and fast recovery
in case of any damage. 
Temporally there are 3 stages 
to make, keep, and improve the system's
resiliency (Fig.~\ref{fig:temporalResil}) pre-, during, and post-event actions are needed.
Extreme events cause a limited or 
complete blackout in the power grid. 
A complete blackout occurs with a low 
probability as a result of multiple 
interrelated contingencies; operators rarely anticipate that.
If pandemics such as COVID-19 coincide 
with extreme weather events, then the 
problem escalates to a whole other level 
where the consequences are extremely high for at least two
reasons. The first reason is that the COVID-19 regulations
and limitations themselves increase difficulties on power grid operators. 
For example, fewer technicians are available to fix failures, while there are more
opportunities for cyberattackers. The second reason is
that hospitals (a very specific example is
ventilators) depend on
electricity~\cite{MichaelLewis19}: ``...
electric power is not just important for the
overall functioning of the facility, it is
critical for direct patient care.''
Automation and ML
can aid in different ways to increase 
the resiliency of networks. 
For example, ML can 
help evaluate the risk of failure and 
consequently with the scheduling of timely
maintenance~\cite{Rudin2012}.
Resiliency has to be implemented 
at multiple levels, though, as different 
incidents threaten the system, e.g., a blizzard vs. human error.

Different approaches that can enhance the 
resiliency of power grid are discussed here.

\subsubsection{Challenges and Opportunities with Integration of Microgrids and DERs}
Although rigorous efforts and suitable solutions to
strengthen the resiliency of the power grid are
continuously evolving, the operator still needs to deal
with challenging conditions due to unprecedented or
adverse events. With the increased deployment of DERs
and microgrids, the reliability and resiliency of the
power grid can significantly be enhanced~\cite{WU2020115254}. 
By reconfiguration of the network, DERs can be used to
serve critical loads when the utility grid or a
centralized generation unit is not available~\cite{GridWise}.
Similar to DERs, microgrids could operate independently
or in connection with the distribution
network~\cite{Sohrab,shahid}. 
Furthermore, a microgrid can be used to 
pick up critical loads outside its boundary 
to increase the resiliency of power 
grids in case of extreme 
events~\cite{Liu2016MicroResil}.
Consequently,
smart grids and distributed renewable energy systems
can be a solution in the future~\cite{MAD39} 
to alleviate problems imposed by pandemics.

Despite the fact that microgrids increase 
the resiliency of power systems, they face 
their own challenges. Machine learning provides
solutions in the microgrid ecosystem as well.
For example, Lin et al.~\cite{8705754} combine neural networks
and support vector machine for state recognition in microgrids.
Their algorithm is used to increase reliability of operation  
by changing the protective settings and the network topology based
on the system’s state.
To learn more about 
these challenges (e.g. physical or cyber threats)
please see Mishra et al.~\cite{MISHRA2020114726}.
Table~\ref{tab:CyberchallengeTablePG} 
provides a 
summary of some of these challenges and some of 
the methodologies developed by researchers 
to overcome the difficulties.
{\small
\begin{table*}[ht]
\setlength{\tabcolsep}{4pt}
 \caption{Examples of Challenges and Opportunities with Integration of Microgrids}\label{tab:CyberchallengeTablePG}
 \begin{tabular}{ l l l l }
\hline
\rowcolor{Gray}
Component & Challenge & Opportunity (Developed Methods) & Refs.\\
\hline
\multirow{1.5}{6em}{System Protection - Cyber Security} & \multirow{2.3}{15em}{Detect and stop cyber attacks \\in wireless sensor networks \\in microgrids with different\\ ownership.} & \multirow{2.3}{19em}{Detecting anomaly based on the
lower and upper bound estimation method to predict
optimal intervals over the smart meter readings at
electric consumers. They make use of the combinatorial
concept of prediction intervals to solve the
instability issues arising from the NNs.} & 
\multirow{2}{2em}{\cite{Abdollah89}} \\ \\ \\ \\ \\ \\ \\
\hline
\multirow{1.5}{6em}{Utility grid connected microgrids} & \multirow{1.5}{15em}{Microgrids with 
distributed\\ generation increase resiliency of power systems 
in case of any faults. Microgrids and using 
\\ML in their routine operations \\gives 
time and space to operators \\to fix the 
damaged components.\\} & 
\multirow{1.5}{18em} {A NN is developed to provide cooperative voltage 
regulation for a microgrid that can 
be applied locally in a privacy preserving way.
} & \cite{kolluri20} \\ \\ \\ \\ \\ \\ \\ \\

& \multirow{2}{15em}{Enhancing protection for \\ modernization of distribution \\system against faults.} &  \multirow{1}{18em}{ A NN and support vector machine is used for state recognition and state diagnosis. 
}  & \cite{lin2019adaptive} \\ \\ \\ 

\hline 

\multirow{1.5}{6em}{Fault detection} & \multirow{1.5}{15em}{Detecting events in each of the distributed stations.} & \multirow{1.5}{18em}{Modified ensemble of bagged decision trees with an added boosting method 
} & \cite{al2018dynamic} \\ \\ \\
\hline
\end{tabular}
\end{table*}
}
\subsubsection{Operational practice for human operators}
Operational practice
guidelines and decision-making tools help the operator to apply the best
plan to minimize the impact of an event, but mainly for known and analyzed events. 
Effective proactive strategies
with quantified measures in terms of
resiliency metric can lead to substantial
improvement in resilient power distribution
grids~\cite{schanda,Venkataramanan2019,Venkataramanan2020},~\cite{gowtham,shikhar}. Training for extreme events will 
be important.
Further, any advancement in automatic decision-making 
tools based on data
processing and analysis will
ease the operator in handling these
events~\cite{MadaniIEEEPSRC}. An
automatic emergency control action plan can be developed
based on previous incidents and can be suggested to operators 
or used as an emergency automated operation.
Distancing employees and scheduling
operating cycles for minimal interactions
with others during health hazards like
COVID-19 stops escalation of impact to
employees, specifically to critical personnel of utility companies,
such as control center operators and field crews.

\subsubsection{Organizational support}
Organizational support plays a 
major role in a resilient grid 
operation, especially given that 
multiple organizations and interests impact grid
operations--federal, 
state, and city governmental 
bodies, FERC, NERC, public 
utility commissions, vendors, 
service providers, technical 
committees, standard development organizations, researchers, and, 
of course, consumers. With proper 
support and guidelines from 
regulation authorities and 
federal energy agencies, 
utility companies could 
accelerate the deployment of 
advanced grid technologies
to perform remote operations 
and increase the level of 
automation with sufficient 
personnel involvement. 
Organizations need to develop policy,
proactive or corrective action plans
against extreme weather (Tsunamis,
Hurricanes, Tornadoes, etc.) and
outbreaks (COVID-19, Ebola, etc.). As the
grid is highly susceptible to cyber threats
during these kinds of events, robust
communication infrastructures and adaptive
action plans must be implemented. During
outbreak events, automated alarm and text
messaging of outages, faults, equipment
failures, and malfunctioning to crew
engineers and field inspectors should be
deployed to reduce the spread of infection
and loss of crew. Automated
routing of field personnel to equipment
maintenance and replacement with minimal
customer or human interaction can ensure
continuous power supply. Moreover, optimal
placement of inventory and crew by
organization greatly diminishes power delivery
interruptions.
Further, mutual coordination and personnel
exchange between organizations operating in
the same region could establish a more resilient and robust grid.
\subsubsection{Redundancy and automation}
Restoration is a key functionality of an advanced 
distribution management system (ADMS) for keeping the lights on after an outage. Redundancy plays a
vital role in the restoration process, and spare
inventory and parallel lines act as major attributes thereof. Outage
management systems (OMS) can identify faulted sections using
information received from trouble calls or field inspectors and
dispatch crews to mend damaged 
equipment~\cite{ChenResil}. An increased
number of multiple power flow paths in the
network permits the load to be
supplied through alternate pathways~\cite{TheFutureOfGrid}. 
Modern grid technologies--automatic metering
infrastructure (AMI), micro-phasor measurement units (microPMU),
intelligent electronic devices (IED), fault indicators,
remote terminal units (RTUs), digital relays and circuit
breakers, data analytical tools--
provide enhanced observability and controllability of
distribution networks~\cite{ChenResil,DepartmentE}.
Field devices can be used to 
automate the power grid
through ML techniques.
For example, predicting the probability of 
failure of a feeder switch can help prevention
of outages. In the case of COVID-19, 
redundancy of spare devices and automation will help in faster restoration.
\subsection{Future ML-based tools and resiliency}\label{MLResiliency}
The need for increasing system resiliency
has been heightened by several recent natural
events and their dramatic impacts. 
But natural disasters are just 
one major 
cause of high-impact outages~\cite{madaniGrip}. 
An example of this kind is 
human-caused outages applicable in 
the COVID-19 case.
Making operators work
longer hours or sequestering them 
to live on-site at power plants will
increase the rate of human error.
With increasing complexity,
the tasks of human operators become even more
challenging.
Unlike a self-driving car, 
a fully automated scenario without human 
intervention is not possible
due to complexity of the power grid.
ML-based tools could help to 
minimize strain on over-burdened
operators and hence minimize the chance of human error. We review
the potential of statistical
learning and recently 
developed techniques to mitigate  potential power
grid challenges as outlined here.

\textbf{ML and natural disasters.}
Gupta et al.~\cite{GuptaSVM} employ
a probabilistic framework and support vector machine to 
predict cascading failure and provide 
early warning in case of any future events to
increase resiliency. Eskandarpour and Khodaei~\cite{Eskandarpour2017ML} 
develop an ML-based prediction method to 
determine the potential outage of power 
grid components in response to an imminent 
hurricane. The decision boundary of the 
classifier is obtained via logistic regression 
to predict the number of failed components and failure duration. 
Predicting the state of a system after 
a hurricane is investigated 
in~\cite{EskandarpourResil} using a support vector 
machine-based model to categorize 
power grid components into two 
classes--damaged and operational. Guikema~\cite{Guikema1111} 
uses regression and data mining techniques 
to estimate the number of poles that 
need to be replaced after a hurricane event.
Glavic et al.~\cite{GlavicRL2017} and 
Huang et al.~\cite{HuangRL2019} explore the
application of reinforcement learning in power
system emergency control dynamics to alleviate
current practices.
In one interesting article,
Maharjan et al.~\cite{8752594}
combine DERs and support vector machines to
improve the resiliency of power grids during
natural disasters. They consider a scenario 
in which different load needs
are categorized to prioritize life threatening cases (e.g.,
a kidney dialysis machine is more important than
a fridge) and ensure power availability for
the most necessary items during outages.

\textbf{ML and cyber attacks.}
Hink et al.~\cite{HinkAdhikari} explore several ML algorithms--such as random 
forests and support vector machine--to classify 
``malicious data and possible cyberattacks'' to help operators in normal 
and emergency situations. 
Foroutan and
Salmasi~\cite{ForoutanDetection} investigate a
Gaussian mixture model for detecting anomalies
and compare their model against some other
models for false data injection. 
Zhang et al.~\cite{ZhangCSI} 
examine a distributed intrusion
detection system embedded at each level of the
smart grid—the home area networks, neighborhood
area networks, and wide area networks using the 
support vector machine
and artificial immune system. 

Protection systems are an essential 
part of the system, yet, they are 
prone to external influences. 
Multiple alarms reported in the control center
could be a result of the faults or failures in
the protection system (unexpected operation). 
Ahmed et al.~\cite{Ahmed2018Cyber} develop a
ML technique to monitor
transmission protection systems and detect
malicious activity using long short-term
memory (LSTM). These data are also
utilized for failure diagnosis. 
Finally, a model to find the root causes of
the observed events assisted by the 
cyber log data from the protection 
devices are developed. Anomaly 
detection in PMUs is further 
discussed in~\cite{Gholami2019} 
to detect the root cause of the 
failure in the transmission 
protection system, and it is 
shown that ensemble models are 
superior to stand-alone methods.

\textbf{ML-based tools to supplement operators.}
ML can help the operators in a
variety of ways. For example, it can be used to
classify the customer ticket texts into different
categories~\cite{Rudin2014} such as 
serious or non-serious events. Natural language
processing and/or recurrent neural networks
can be used to analyze customers' trouble
calls to pinpoint the location of the faulty
equipment as opposed to operators taking the
calls. By automating this process,
interaction among operators will be minimized as needed 
in the case of COVID-19.

For dispatchers and operators to manage a 
power grid correctly and efficiently, they 
need to understand the dynamics and the current state of the grid. Using alarms generated by
SCADA is one of their tools. The rate of alarm 
generation can be reduced significantly (98\%) by using
an intelligent alarm processing system, which is
a tremendous improvement for operators and dispatchers~\cite{BaranovicIAP}. 
Interventions like these are critical, especially in emergencies, 
since the long work shifts and high levels of 
stress that occur over times of emergency 
(such as COVID-19) make operators even more prone to mistakes~\cite{PandeyGM}. Furthermore, ML can be utilized to classify the alarms
and hence reduce the workload of operators. 
Employing ML can help to identify multiple events associated with a set of reported alarms, consequently reducing the operators' cognitive burden.
\subsection{A case study for Resiliency Management Tool with NLP}\label{caseStudy}
\begin{figure*}[ht]
  \centering
  \captionsetup{justification=centering}
  \includegraphics[width=.7\textwidth]{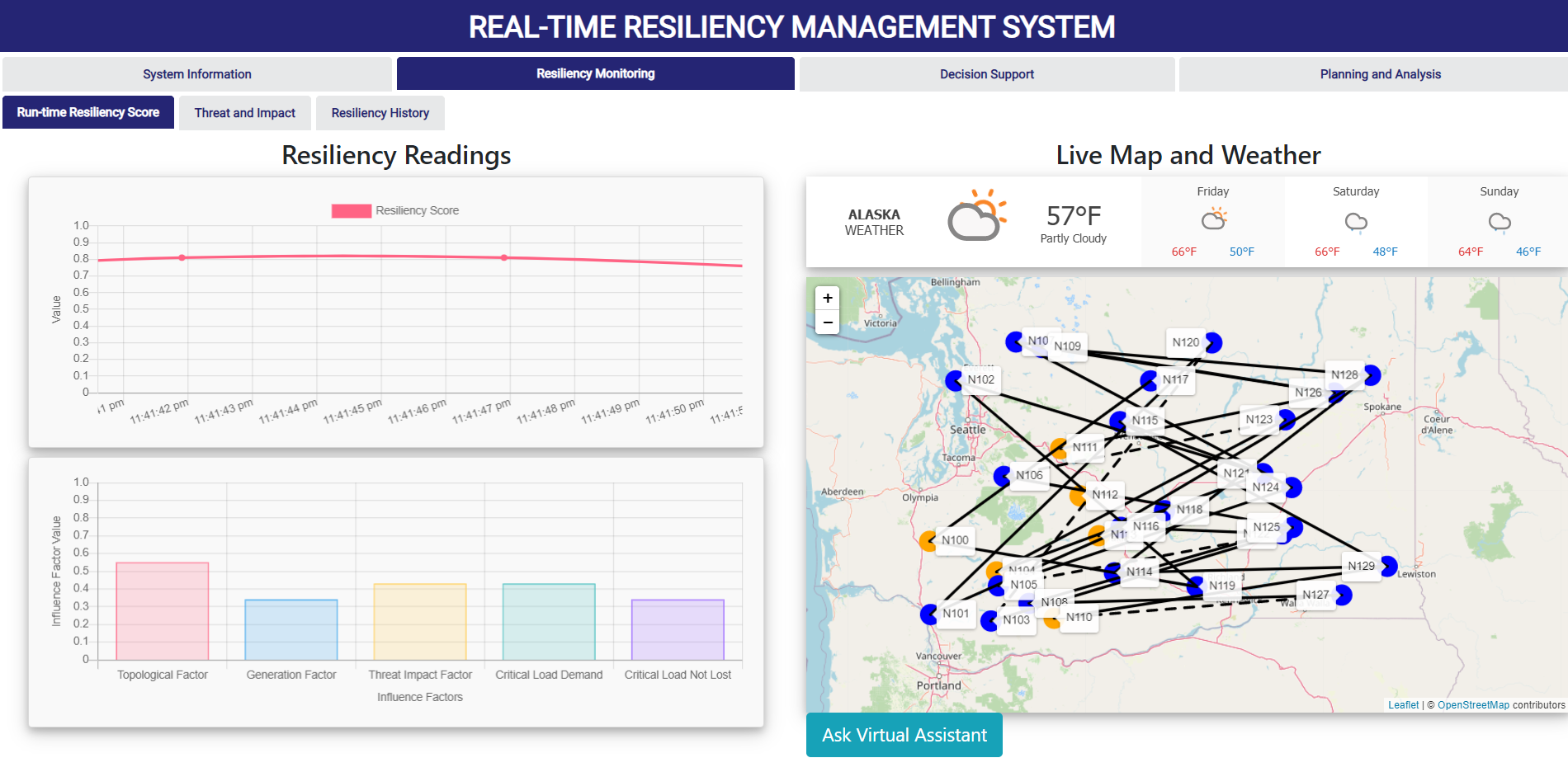}
  \caption{RT-RMT tool view: Normal operating condition.}
  \label{fig:SanjeevFigure1}
\end{figure*}

A real-time resiliency 
management tool (RT-RMT) is a unique 
tool developed by the authors'
team. It provides the resiliency score
of any power system network by
considering the various factors such as power
generation, critical and non-critical 
loads, threat type,
and impact model including available 
backup resources,
inventory, field crew, and inspection 
team, and weather information \cite{gowtham2020}. 
This tool also encompasses
geospatial information of the power grid, and  
assesses  an infrastructure condition
 to compute resilience 
scores during the propagation stages
of natural threats. 
Three stages of
events, pre-, during, and 
post-event, are considered to provide resiliency
scores for each stage by taking relevant influencing factors into account.
The real-time resiliency score 
is computed every 15 minutes using the
information acquired through communication channels 
from different
devices (e.g., SCADA, PLC, RTUs) in the distribution 
system. It can be deployed along with
a commercial distribution management system software. RT-RMT
provides real-time operational topology or configuration
on a live tracking
map shown in Fig.~\ref{fig:SanjeevFigure1}. 
Further, it displays the
connection between different nodes 
of the network and routing over
the GIS layer. It can be adapted to 
analyze the impact of health
hazards (e.g., COVID-19)  and can effectively minimize 
its influence on the electrical
distribution system operations.
\begin{figure*}[ht]
  \centering
  \captionsetup{justification=centering}
  \includegraphics[width=.7\textwidth]{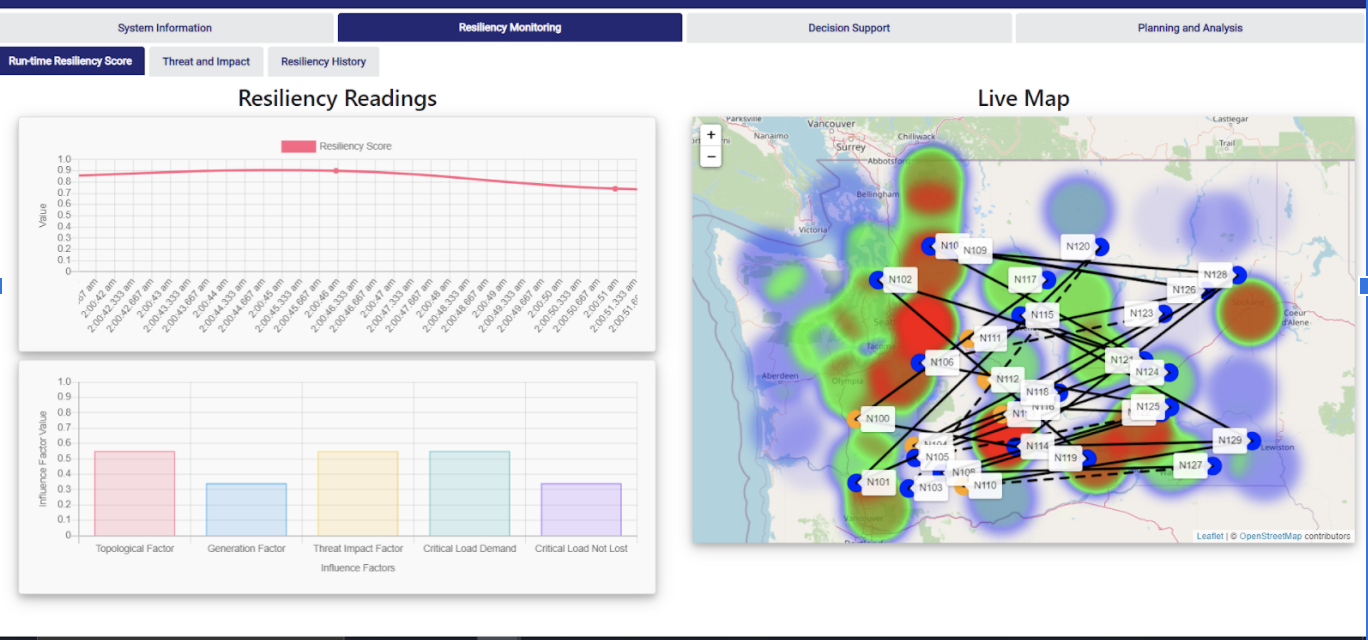}
  \caption{RT-RMT tool view: Spread of COVID-19 cases across Washington state with blue being lowest and red being the highest number of cases.}
  \label{fig:SanjeevFigure2}
\end{figure*}
RT-RMT tool has four main windows; system 
information, resilience monitoring, decision support, 
and planning analysis. System information portrays  
physical assets ( number of buses, transformers, fault indicators,
protection devices, other equipment), and their
ratings along with deployed location and current 
operating status. Resilience monitoring captures 
real-time resiliency score for the provided distribution system. This 
value will change subject to operating
conditions (i.e. with variations in the generation, 
loads connected, etc.). The decision support on RT-RMT helps the operator take appropriate actions suggested by the tool to boost the system's resiliency, including pre-, during, and post-event requirements as the event progresses through different stages. Planning and analysis of tool emphasizes enhancing the resilience of system operation with system upgrades like adding distributed energy resources, increased energy reserve, fast communication technologies, sensors, etc.

Two use cases are developed to evaluate the performance of the RT-RMT tool. The first use case focuses on the normal operation of the distribution system without any event. The evaluated resiliency score for a given operating condition is near to 0.9 as shown in Fig.~\ref{fig:SanjeevFigure1}. 
The second use case includes the 
COVID-19 pandemic. It shows how RT-RMT will be adapted 
to help the operators address the problems with
minimal impact on the electrical power supply and all 
the consumers. The data used in this scenario are 
obtained from John Hopkins' Coronavirus
dataset~\cite{JHUData} and are anonymized to avoid 
data privacy issues. Different regions with different
intensities of COVID-19 cases are shown in different
colors in Fig. \ref{fig:SanjeevFigure2} over the
distribution system in a live tracking map. This tool
informs the operators (visually) about areas with high
cases of COVID-19. Consequently, they can decide to 
change any scheduled maintenance plan or repair a 
failed equipment plan as needed. The operator may 
address the issues in regions with a low number of 
COVID cases, first with precautions. For example, by
sending limited crew per vehicle. RT-RMT also 
optimizes the route to avoid the paths going through
regions with a high number of cases to
protect crew members against infection. 
The operator could also plan to send crew 
members at once in a day along with
PPE kits to ensure minimal contact of humans. Further, effective reconfiguration 
of the network can also be performed in 
regions with high positive cases
to restore the power after an outage or 
equipment failure so that electrical supply is 
still maintained without field crew inspection 
and replacement of faulty equipment.
Detailed explanation is provided in the section
\ref{sec:sanjcrew}.

\begin{figure*}[ht]
  \centering
  \captionsetup{justification=centering}
  \includegraphics[width=.7\textwidth]{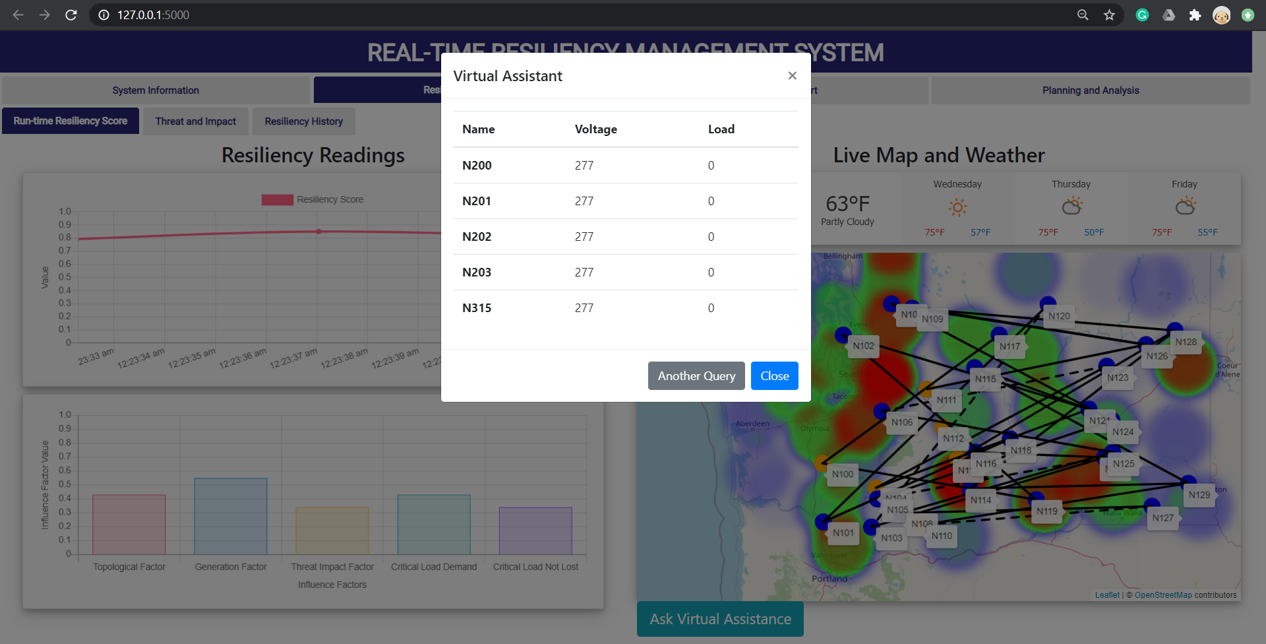}
  \caption{Virtual assistant demonstration on RT-RMT.}
  \label{fig:SanjeevFigure3}
\end{figure*}

Digital query assistant is developed using 
natural language processing and plugged into the 
RT-RMT tool with the name ``Ask Virtual Assistant'' 
as shown in Fig.~\ref{fig:SanjeevFigure1}. 
This can be used to dig the required information 
from the database where high data volumes from the
power system network are stored. A virtual 
assistant will ease operator
searches for necessary information during normal 
operating scenarios as well as emergency situations 
(such as outages, weather-related events, pandemics). 
The virtual assistant also lessens the cognitive 
burden on operators, reducing human errors in
control decisions. A typical demonstration is 
showcased here by asking the query: 
``what node has the lowest voltage''? Results 
appear within a small window over the main screen 
as shown in Fig.~\ref{fig:SanjeevFigure3}.
This can be further enhanced with a virtual 
voice assistant--similar to Google and Apple
applications--which helps the operator to locate 
necessary information.
\subsubsection{Use case of RT-RMT for optimal crew routing during COVID-19 pandemic} \label{sec:sanjcrew}
\begin{figure*}[ht]
  \centering
  \captionsetup{justification=centering}
  \includegraphics[width=1\textwidth]{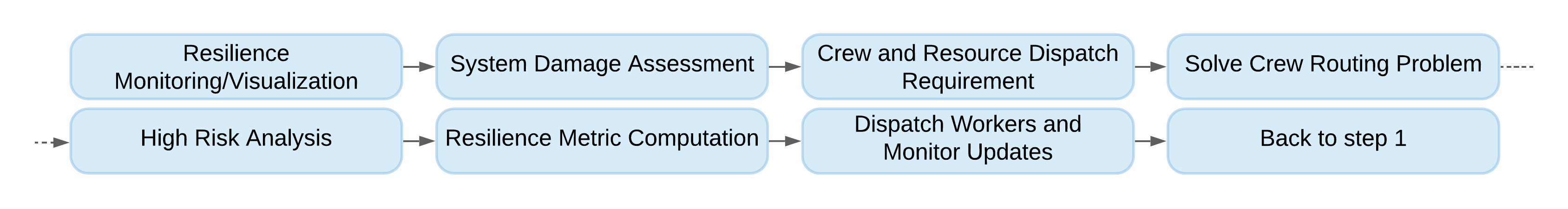} 
  \caption{Safe crew-dispatch algorithm.}
  \label{fig:Sanjeev_crew1}
\end{figure*}

The distribution test system is a 45-node 
simplified distribution network that is remote and 
with a large DERs supporting most of the operation, 
as shown in Fig.~\ref{fig:SanjeevFigure1}. The 
dataset for the COVID-19 is derived from the 
CORD-19 (COVID-19 Open Research Dataset) 
presented by Wang et al.~\cite{wang2020cord}. 
In normal operating conditions, the RT-RMT 
only supports operator actions while tracking 
the resilience performance of the system. 
In the 
case of the COVID-19 pandemic, the authors 
developed a safe crew-routing algorithm, 
presented in Fig.~\ref{fig:Sanjeev_crew1} 
that solves the crew routing problem for 
system repairs. It then filters the solution 
set for the crew route and schedules a path 
with the least risk of contracting COVID-19. 
The filtered solution set is then subject 
to a comparative analysis through the 
resilience metrics to identify the most 
resilient crew route and is visualized 
in the RT-RMT live map for operator sign-off. 
The resilience metric that captures the 
resilience performance of the system after 
the extreme event has ended gathers the 
resilience indicators that inform the 
rapidity of recovery for each restoration 
plan that is proposed by the operator.

A vector of recover resilience indicators is formed ($\mathbf{R} = [T_r,~C_r,~\tau ,~CL_r,~SO]$)
for each path. The indicator $T_r$ represents the total 
repair time, $C_r$ represents the cost of repairs, 
$\tau$ represents the topological resilience coefficient that
is based on graph-theoretical spectral metrics~\cite{schanda},
$CL_r$ represents the number of critical loads restored in the
process, and SO is the number of switching operations. For
each path, the vector $\mathbf{R}$ is computed, and the
analytical hierarchical process~\cite{gowtham} 
is applied to find the composite resilience score.
Resilience is maximally enabled when the 
operator selects a path with the highest
resilience metric score.

\begin{figure*}[ht]
  \centering
  \captionsetup{justification=centering}
  \includegraphics[width=.7\textwidth]{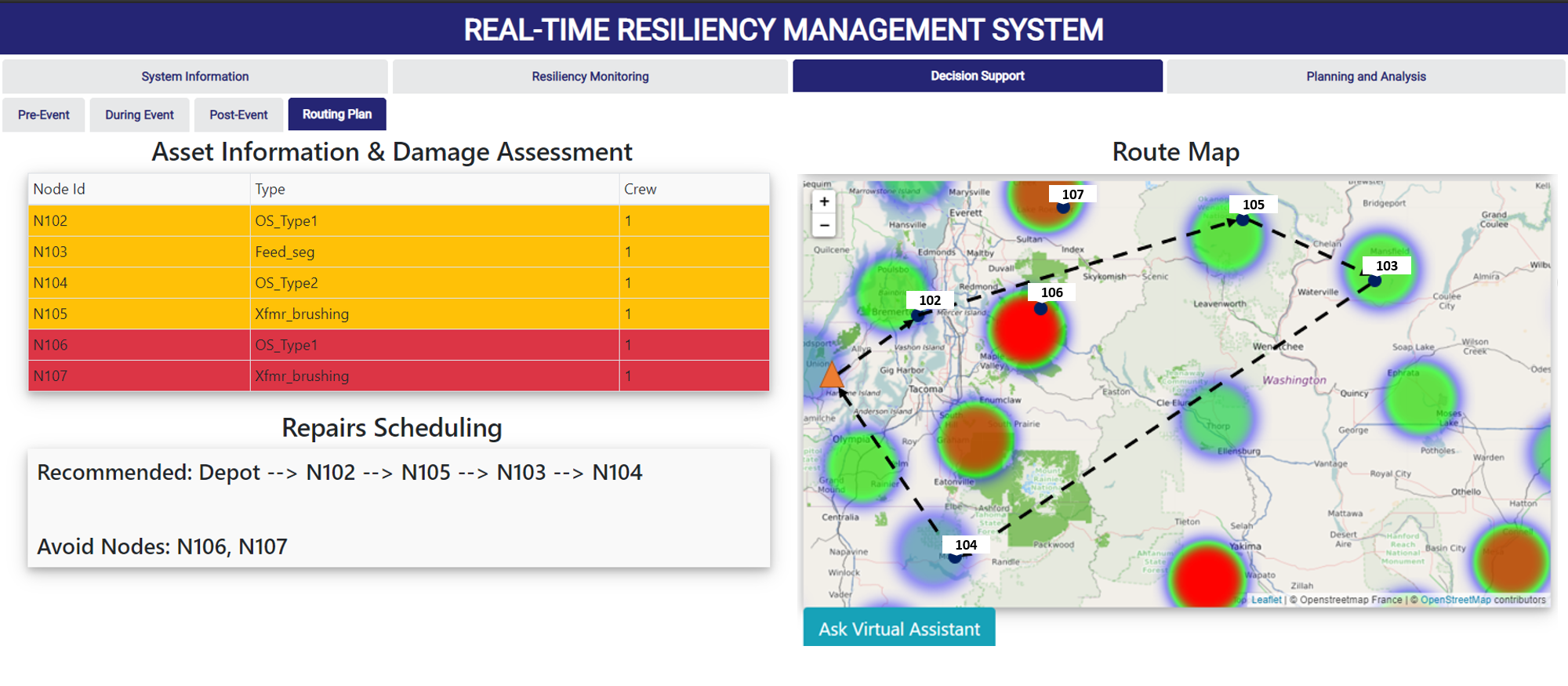}
  \caption{Screen capture of RT-RMT showing the asset information with repair instructions, scheduling with no-go zones, and the routing on the map with  crew-routing algorithm.}
  \label{fig:Sanjeev_crew2}
\end{figure*}

A simple eight-node repair scenario is 
analyzed in the test system from
Fig.~\ref{fig:SanjeevFigure2}. 
Once the RT-RMT obtains the damage assessment
scenario, the resilience computation engine applies
the routing algorithm to get the lowest weight route
with the GIS, COVID-19 data, repair information, and
scheduling requirements so that exposure of field
crew to COVID-19 hotspots is avoided.
Figure~\ref{fig:SanjeevFigure2} shows the main
operator screen with the resilience metrics
calculations. The position of each node and relative
proximity with COVID-19 hotspots are displayed on the
tool UI (Fig.~\ref{fig:Sanjeev_crew2}). Three routes
are analyzed using the resilience metrics analysis,
and the most resilient option that is also the
safest is chosen. The safest route with 
avoided zones is shown 
in Fig.~\ref{fig:Sanjeev_crew2}.
\section{Future Work and Path Forward}\label{futureDir}
The data needed for statistical learning can be divided into two parts: environmental data and power system data~\cite{7105972}. The ecological situation (e.g., precipitation, soil moisture, wind speed, etc.) is not affected by the pandemic. Some parts of power systems data, such as component failures or the nature of cyberattacks, are not also affected by COVID-19. Hence, the rarity of COVID-19 affects neither the power system data nor the environmental data. Consequently, ML-approaches can be beneficial in pandemic situations. One challenge that needs to be addressed is finding efficient approaches to deal with sudden behavior changes in energy consumption patterns in terms of short-term forecasting.

As reported by many resources such
as~\cite{Paaso}, many control center experts
had to work long hours, which leads to 
fatigue (as well as other problems), 
potentially resulting in human error. 
Implementing ML approaches can reduce
the cognitive workload of humans in such cases.

Statistical learning techniques, e.g., natural language processing, can be employed to analyze incoming calls of customers to pinpoint the location of failures in the power grid. Investing in the implementation of ML approaches is worthwhile to help customer operations and office staff by automatically detecting failed devices/locations, even without taking calls.

The IT staff were forced to work remotely which created opportunities for cyberattackers.
Implementing ML approaches to detect suspicious
changes in the power grid to protect against
cyberattacks is beneficial to IT experts.
Opinion dynamics have 
been studied for about 50 years; however, 
they have attracted a lot of attention
recently~\cite{recentOpin,classical}. 
In the past few years, opinion 
dynamics are being
used to thwart attacks on
networks~\cite{shang2019consensus,rubio2019enhancing,8304395}. 
Therefore,
combining ML and opinion
dynamics is a potential future technique that could be
applied to power grids to combat cyberattacks.

Utilizing topological data analytics 
in anomaly detection in power grid
studies are limited. Two examples of 
topological data analytics
are~\cite{ChazalTDA,KamruzzamanTDA}. 
Topological data analytics
in~\cite{KamruzzamanTDA} is used to identify
the divergent subpopulation in 
phenomics data
throughout the growth period of plants.
Such an approach might potentially 
detect the root cause of failures and early
warning when equipment starts to behave
abnormally. Moreover, Chen et al.~\cite{8587547}
have designed an algorithm that reveals 
ML-algorithms' 
(such as Recurrent Neural Networks) 
vulnerability to disturbance of input data. 
Attackers might manipulate the 
input data in order
to affect the output of ML-algorithms. Chen et al.
argue that current algorithms cannot detect 
such disturbances.
Therefore, it would be worthwhile to investigate the
approach of~\cite{KamruzzamanTDA} to detect 
such disturbances.
Furthermore, ML is also being explored for 
\textit{cyber situational awareness}~\cite{FrankeUlrik} and needs
more development in the future.

The existence of large amounts of data (e.g., collected by PMUs) is both a gift and a curse.
From a hardware point of view, 
there is a need to either build an
infrastructure that can handle that amount of
data or long-term collaborations 
with Micorosft's Azure or Amazon's AWS 
need to be established. From a software point 
of view, there are packages that implement 
some ML techniques to
enable dealing with data at scale. 
However, there are some challenges:
some of the implemented techniques are very 
costly in terms of storage 
and computations (e.g., iterative methods such as
neural networks), 
or not all ML techniques are implemented 
in a given software/library. Zheng and
Degnino~\cite{7004327} provide a detailed 
survey of available resources and challenges in 
this regard.

Finally, we remind the reader of a lack of 
contingency plans for human aspect of power
systems. COVID-19 revealed the vulnerability 
of power systems being implicitly affected 
as a result of human operators'
vulnerability. Thus, there is an important gap
in our knowledge that needs to be addressed.
Filling this gap will need
financial and organizational support but will
be beneficial long term.
Regardless of the dearth of data 
on operator vulnerability,
resiliency, and/or contingency 
plans for human assets,
the implementation of data-driven approaches
discussed in this paper will
increase the resiliency of the 
power grids so that normal operations
can continue smoothly under circumstances such
as COVID-19. The presence of data-driven techniques
can be considered a prevention strategy for
any interruption--e.g., the decrease in working hours
of operators or the reduction in the number of human
operators on site.

Similar to protection plans for
physical assets 
(e.g., water barriers for substations), there
is a need to build similar protection plans for human assets.
Similar concepts can
be carried out for humans--e.g., 
building onsite amenities for
humans to stay for long shifts.

\section{Conclusions}\label{conclusionSec}
As observed over the last several years, the frequency 
and intensity of natural disasters and extreme events 
and their associated impact on the power grid have been increasing. 
Pandemics such as COVID-19 add to these challenges and push 
organizations to their limits. Driven by smart grid
investment, the power grid is moving towards massive
sensor deployment and automation. Enhanced digitalization has
resulted in the collection of so much data that 
it is beyond the capacity of human operators 
to analyze, use, and factor it into timely decisions 
when responding to extreme events. 

In this paper, the impact of COVID-19 on power grid
operations has been analyzed. Actions being taken by
operators/organizations to minimize the effects of
COVID-19 has been discussed in detail. Solutions have
been suggested for deploying recently developed
tools and concepts in ML and AI that will increase the
resiliency of power systems in general and in extreme
scenarios such as the COVID-19 pandemic. Suggested tools
can assist operators in taking control actions, allowing
collaborative machine-human interactions while using increased
data integration from diverse sources in an
intelligent manner. Data-driven ML and AI could benefit operators by helping with the systems' resiliency, decision making, and disaster management in smarter ways. For example, predicting disaster impact,  anomaly detection, and overlapping the COVID-19 hotspots with electricity service maps helps control room operators restore the grid fast and effectively by considering all safety measures.
A sample case study of data-driven RT-RMT tool is presented to demonstrate the usefulness of advanced tools to system operator in case of pandemic and for enhanced resiliency.
\phantomsection





\bibliographystyle{IEEEtran}

\typeout{}
\Urlmuskip=0mu plus 1mu\relax
\bibliography{covid.bib}

\end{document}